%% file: main.tex
\documentclass[letterpaper, 10 pt, conference]{ieeeconf}

\input{notation}

\input{usepackage}

\bibliography{books, papers}

\IEEEoverridecommandlockouts
\title{\LARGE \bf Multi-Hypothesis Interactions in\\ Game-Theoretic Motion Planning}

\author{
Forrest Laine, David Fridovich-Keil, Chih-Yuan Chiu, and Claire Tomlin
\thanks{
F Laine, C-Y Chiu, and C Tomlin are with the Department of Electrical Engineering \& Computer Sciences, UC Berkeley. D Fridovich-Keil is with the Department of Aeronautics and Astronautics, Stanford University. Correspondence to \href{mailto:forrest.laine@eecs.berkeley.edu}{\tt \small{forrest.laine@eecs.berkeley.edu}}.}%
\thanks{
This research is supported by an NSF CAREER award, the Air Force Office of Scientific Research (AFOSR), NSF's CPS FORCES and VeHICaL projects, the UC-Philippine-California Advanced Research Institute, the ONR MURI Embedded Humans, a DARPA Assured Autonomy grant, and the SRC CONIX Center.}
}

\begin{document}

\maketitle
\thispagestyle{empty}
\pagestyle{empty}

\begin{abstract}
We present a novel method for handling uncertainty about the intentions of non-ego players in dynamic games, with application to motion planning for autonomous vehicles. Equilibria in these games explicitly account for interaction among other agents in the environment, such as drivers and pedestrians. Our method models the uncertainty about the intention of other agents by constructing multiple hypotheses about the objectives and constraints of other agents in the scene. For each candidate hypothesis, we associate a Bernoulli random variable representing the probability of that hypothesis, which may or may not be independent of the probability of other hypotheses. We leverage constraint asymmetries and feedback information patterns to incorporate the probabilities of hypotheses in a natural way. Specifically, increasing the probability associated with a given hypothesis from $0$ to $1$ shifts the responsibility of collision avoidance from the hypothesized agent to the ego agent. This method allows the generation of interactive trajectories for the ego agent, where the level of assertiveness or caution that the ego exhibits is directly related to the easy-to-model uncertainty it maintains about the scene. 

\end{abstract}

\input{intro}

\input{related_work}

\input{preliminaries}

\input{methods}

\input{results}

\input{conclusion}



\balance
\printbibliography

\end{document}

%% file: notation.tex
\newcommand{\player}[1]{$P^{#1}$}

\newtheorem{definition}{Definition}

%% file: usepackage.tex
\usepackage{graphicx}
\usepackage{adjustbox}
\usepackage{wrapfig}
\usepackage[labelfont=bf,figurename=Fig.,font=small]{caption}
\usepackage{subcaption}
\usepackage{lipsum}
\usepackage{amssymb}
\usepackage{amsmath}
\usepackage{siunitx}
\usepackage{mathtools}
\usepackage{xcolor}
\usepackage{hyperref}
\usepackage[capitalize]{cleveref}
\usepackage[ruled,vlined,linesnumbered]{algorithm2e}
\usepackage[noend]{algpseudocode}
\usepackage{ifthen}
\usepackage{dsfont}
\usepackage[backend=biber,style=numeric-comp,sorting=none]{biblatex}
\usepackage{balance}
\usepackage{dirtytalk}

\usepackage{times}

%% file: intro.tex
\section{Introduction}
\label{sec:intro}

Trajectory planning for autonomous agents often proceeds in a receding time horizon. When operating in the presence of other agents, as in autonomous driving, these plans must reason about a predicted future and account for properties such collision-avoidance, by using predictions of other agents' future decisions.
A difficulty of this approach is that the resultant motion plans of the ``ego'' vehicle are incapable of reasoning about the reactions of other agents to its own decisions. A canonical example of this problem is that of an autonomous vehicle unable to merge into dense traffic because it cannot reason that other agents will make space.

Game-theoretic motion planners account for these reactions by handling planning and prediction jointly. That is, they encode the intentions of other agents in the scene as optimization problems that they are each trying to solve, and generate an equilibrium solution. This equilibrium consists of a set of interacting trajectories of all the agents; the ego agent can then follow its own equilibrium trajectory.


\begin{figure}
    \centering
    \includegraphics[width=\columnwidth]{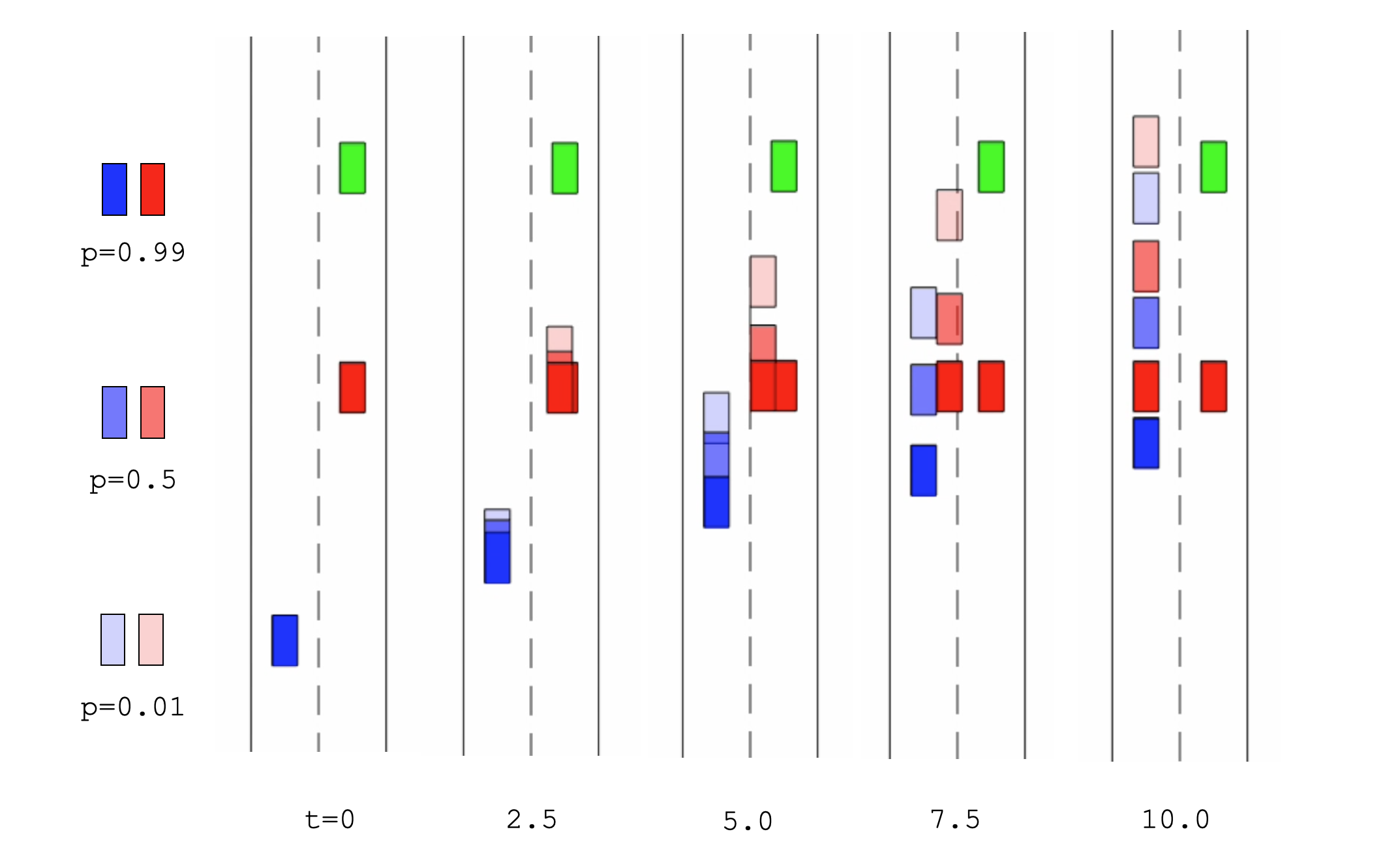}
        \caption{The fast-moving ego vehicle (blue) believes that the slow-moving (red) agent will change lanes with   probability $p$. Our method constructs and solves a dynamic game involving both lane-changing and non-lane-changing versions of the red agent. Depicted here are three different solutions to this game, corresponding to different probabilities associated to the two hypotheses. When certain that the red agent will change lanes ($p\approx1$), the ego agent takes full responsibility for avoiding collision with the lane-changing version of the red agent, allowing it to change lanes unimpeded. When certain the red agent will stay in-lane ($p \approx 0$), the lane-changing agent is required to take full responsibility for collision-avoidance with ego, which allows ego to pass at its original speed. Probabilities in the range $0 < p < 1$ result in trajectories which qualitatively interpolate between these two behaviors. Note that the non-lane-changing version of the red agent is unaffected by the actions of the ego agent for all values of $p$ in this case.}
    \label{fig:front}
\end{figure}

A serious drawback of game-theoretic planning is that it assumes that the intentions of other agents are known to the ego, and that the agents act rationally with respect to those intentions. In practice, outside the context of game-theoretic planning, these objectives and intents can be estimated online and planned for as contingencies \cite{hardy2013contingency}. In this paper, we address this challenge in the context of game-theoretic planning and present a multi-hypothesis framework that accounts for this uncertainty and associated coupling. 

In particular, our approach assumes the ego agent maintains a discrete set of hypotheses regarding the intentions of other agents. These hypotheses have probabilities associated with them, and model uncertainties about intentions such as potential lane changes, nominal speeds, aversion to acceleration, etc.
Given uncertainty of this nature, we propose constructing distinct optimization problems for each agent, which when optimized over, result in trajectories which are qualitatively representative of the different hypotheses. Replicas of each agent (one for each hypothesis) are created, and are assigned to the corresponding optimization problem. We introduce a novel way of formulating the resultant dynamic game which allows the ego agent to reason about interactions with the replica agents based on the probability of their corresponding hypothesis. This formulation is key to reasoning about collision-avoidance constraints in a probabilistic way. 

Modeling the problem in this way results in pragmatic approach to this general class of motion-planning problems that autonomous vehicles must solve. Our method generates interactive trajectories among multiple agents while explicitly accounting for uncertainty about others' intentions. We hope that our method will enable more practical use of game-theoretic planning for autonomous vehicles.

This paper is outlined as follows. \cref{sec:related_work} discusses other methods for multi-agent planning, including game-theoretic methods as well as methods using derivatives of rational choice models,
which account for the uncertainty of other agents. \cref{sec:preliminaries} reviews concepts related to game-theoretic planning needed to formalize our method, which is presented in \cref{sec:methods}. In \cref{sec:results}, we present empirical results of our method on various realistic driving scenarios, and finally, we make concluding remarks in \cref{sec:conclusion}.

%% file: related_work.tex
\section{Related Work}
\label{sec:related_work}

The method we present builds directly on recent work on specialized game-theoretic planning algorithms and their use for autonomous systems. 

Examples of game theoretic planners include \cite{cleac2019algames}, in which an augmented Lagrangian method is used to search for stationary points with respect to the first-order necessary conditions for the open-loop generalized Nash equilibrium of trajectory games. The authors in \cite{di2019newton} and \cite{di2020local} similarly present a method based on differential dynamic programming which also searches for stationary points of an Open-loop generalized Nash equilibrium. 

The authors in \cite{fridovich2019efficient} propose a method similar to that in \cite{di2019newton}, which approximately computes stationary points for a Feedback Nash equilibrium. The difference between Open-loop and Feedback Nash equilibrium is discussed, for example, in \cite{basar1999dynamic}. The authors in \cite{schwarting2019stochastic} also present a method to solve for approximate Feedback Nash Equilibria of trajectory games, although in the context of belief-space games. The work in \cite{schwarting2019stochastic} accounts for uncertainty in the game caused by partial and noisy observations, which is a different form of uncertainty than what we consider here.

All the above-listed methods can be seen as algorithms for solving Generalized Nash Equilibrium problems (GNEPs) in different contexts, with advantages tailored for those contexts. A survey on GNEPs is given in \cite{facchinei2010generalized}, and a study of Newton-type methods for solving GNEPs is given in \cite{facchinei2009generalized}. 


It is important to note that there are a vast number of works which have considered planning 
under
uncertainty regarding the behavior of other agents in an environment. These methods do not work in a game-theoretic context, and therefore these existing works account for uncertainty in state and input, rather than in objectives and constraints. Regarding autonomous vehicle navigation, \cite{xu2014motion} describes planning in the context of uncertain (yet static) predictions for other agents in the scene. \cite{aoude2013probabilistically} similarly present a real-time method for chance-constrained collision avoidance problems, while \cite{hardy2013contingency} uses optimization-based path planning to establish probabilistic collision avoidance guarantees for autonomous vehicles. Meanwhile, \cite{Sadigh2016information, sadigh2016planning, peters2020inference, ziebart2009planning} describe methods for human activity prediction, in which safety guarantees are established for autonomous vehicles by collecting information to infer the intent of the human agents in the environment. Additionally, distributionally-robust optimal control methods such as \cite{van2015distributionally, nishimura2020rat} relax the common Gaussian uncertainty assumption in chance-constrained optimization.

%% file: preliminaries.tex
\section{Preliminaries}
\label{sec:preliminaries}

This section presents concepts and definitions in game-theoretic motion planning that are necessary for the development of our proposed method in the following sections.


\subsection{Generalized Nash Equilibria}
Consider a game played among $N$ agents, each of which controls private decision variables denoted by $y^i \in \mathcal{Y}^i$. Let there also be a shared decision variable common to all agents, denoted by $z \in \mathcal{Z}$. Let each agent $i$ attempt to optimize the following problem:
\begin{subequations}
\label{eq:generalized_nash}
\begin{align}
    \min_{y^i,z} & \ \  l^i(z, y^1,...,y^N) \\ 
    \text{s.t} \ \ & f(z,y^1,...,y^N) = 0 \\
    & g^i(z,y^1,...,y^N) \geq 0 \\ 
    & h^i(z,y^1,...,y^N) = 0
\end{align}
\end{subequations}
Here the constraint $f(z,y^1,...,y^N) = 0$ is common to all agents, and the dimension of $f$ is equal to the dimension of $z$. We wish to solve for a Generalized Nash Equilibrium of the collection of $N$ problems given in the form \cref{eq:generalized_nash}.
\vspace{.3cm}
\begin{definition} 
A Generalized Nash Equilibrium \cite{facchinei2010generalized} for the collection of $N$ problems \cref{eq:generalized_nash} is a collection of variables $z^*, y^{1*},...,y^{N*}$ such that for all $i$,
\begin{equation}
    \label{eq:nash_condition_objective}
    \begin{aligned}
    l^i(z,y^{1*},...,y^{(i-1)*},y^i,y^{(i+1)*},...,&y^{N*}) \geq \\ l^i(z^*,&y^{1*},...,y^{N*}),
    \end{aligned}
\end{equation}
for all $y^i$ and $z$ satisfying 
\begin{subequations}
\label{eq:nash_condition_constraints}
\begin{align}
    & f(z,y^{1*},...,y^{(i-1)*},y^i,y^{(i+1)*},...,y^{N*}) = 0 \\
    & g^i(z,y^{1*},...,y^{(i-1)*},y^i,y^{(i+1)*},...,y^{N*}) \geq 0 \\ 
    & h^i(z,y^{1*},...,y^{(i-1)*},y^i,y^{(i+1)*},...,y^{N*}) = 0
\end{align}
\end{subequations}
\end{definition}
\vspace{.3cm}
A Local Generalized Nash Equilibrium for the game given by \cref{eq:generalized_nash} is defined analogously, where we only require the condition \cref{eq:nash_condition_objective} to hold for all $y^i$, $z$ satisfying \cref{eq:nash_condition_constraints} and in a local neighborhood of the values $y^{i*}$ and $z^*$. A Generalized Nash Equilibrium is a Nash Equilibrium in which the constraints imposed on each player depend in general on the decision variables of other players.

Numerical methods for computing Generalized Nash Equilibria often compute stationary points of the first-order necessary conditions of optimality of the game, which are formed by concatenating the KKT conditions of each agent's optimization problem (\ref{eq:generalized_nash}). See  \cite{facchinei2009generalized}, for example, for details on Newton-type methods for finding such stationary points. 

\subsection{Dynamic Feedback Games}
\label{subsec:dynamic_feedback}

This work focuses on a special case of the games defined by \cref{eq:generalized_nash}, which are called dynamic games. We use the term dynamic game to refer to a scenario comprised of $N$ agents acting in a discrete-time, continuous-space environment, for which the global state at some time-step $t$ can be represented by the variable $x_t \in \mathcal{X}$, where $\mathcal{X}$ defines the state-space for the environment (often $\mathbb{R}^n$ for $n$-dimensional state-spaces). Often, the state-space $\mathcal{X}$ is the product space of the state-spaces for individual agents, i.e. $\mathcal{X} := \mathcal{X}^1 \times ... \times \mathcal{X}^N$. When this is the case, $x_t = (x_t^1,\dots,x_t^N)$. The agents influence the state of the environment by applying private control variables, denoted as $u_t^i \in \mathcal{U}^i$ for agent $i\in \{1,\dots,N\}$ at time-step $t$. We denote the dimension of $\mathcal{U}^i$ by $m^i$ (both $\mathcal{U}^i$ and $m^i$ may differ across agents), and use the shorthand $u_t := (u_t^1, \hdots, u_t^N)$ and $u_t^{-i} := (u_t^1,\dots,u_t^{i-1},u_t^{i+1},\dots,u_t^N)$. The state evolution of the system over discrete time-steps is given by the dynamic update
\begin{equation}
    x_{t+1} = f(x_t, u_t^1, \dots, u_t^N) = f(x_t,u_t)
\end{equation}

We consider finite-horizon games of discrete time-step length $T$, and assume without loss of generality that games start at $t=0$ from a known state $\hat{x}_0$. The objective and constraints imposed on the actions of the agents in the game are also assumed known. Uncertainty in the initial state of the game, as well as in the objective and constraints imposed on other agents, is accounted for in \cref{sec:methods}.

The Dynamic Feedback Game of consideration is recursively-defined by the sub-game optimization problems for each player $i$, starting at time $s \in \{0,...,T-1\}$ from state $\hat{x}_s$:

\begin{subequations}
\label{eq:agent_fb_optimization}
\begin{align}
    V_s^i(\hat{x}_s) := &\min_{\substack{x_s,\hdots,x_{T} \\ u_s^i,u_{s+1},\hdots,u_{T-1}}} \  \sum_{t=s}^{T-1} l^i(x_t,u_t) + l_T^i(x_T) \label{eq:obj} \\ 
     \text{s.t.} & \ \ x_s - \hat{x}_s = 0, \\
     & \ \ x_{t+1} - f(x_t,u_t) = 0, \ \ s \leq t \leq T-1\\
     & \ \ h_t^i(x_t,u_t^i) = 0, \ \ s \leq t \leq T-1 \label{eq:eq_t} \\
     & \ \ g_t^i(x_t,u_t^i) \geq 0, \ \ s \leq t \leq T-1 \label{eq:ineq_t} \\
     & \ \ h_T^i(x_T) = 0 \label{eq:eq_T} \\
     & \ \ g_T^i(x_T) \geq 0 \label{eq:ineq_T} \\
     & \ \ u_t^{-i} - \pi_t^{-i}(x_t) = 0, \ \ s+1 \leq t \leq T-1 \label{eq:policy_constraints}
\end{align}
\end{subequations}

The cost functions $l^i(\cdot,\cdot)$ and $l_T^i(\cdot)$ are assumed to be continuous and twice-differentiable, and in general may depend on the control variables of other agents, $u_t^{-i}$.

The equality and inequality constraints imposed on each agent are unique to that particular agent in general, but the dynamic constraints (and initial condition) are common to all agents. The dimension of the constraints $h_t^i$ and $g_t^i$ may vary at every time-step $t$, including potentially having dimension $0$ (representing no constraint). We assume these constraint functions are twice-differentiable in their arguments. 

Here, both the shared state variables and the control variables for agent $i$ are treated as decision variables in the optimization, as are the control variables of other players $u_t^{-i}$ for time $t>s$, where they are constrained by a feedback policy $\pi_t^{-i}(x)$. The polices $\pi_t^{-i}(x)$ are defined implicitly to yield the controls $u_t^{-i}$ which form a Generalized Nash Equilibrium for the set of problems $\{V_t^1(x),...,V_t^N(x)\}$.

Solving for the Generalized Nash Equilibrium solution of the sub-game starting at $s=0$ defines the equilibrium for the entire Dynamic Feedback game of interest. This is referred to as a Feedback Nash Equilibrium. Note that this equilibrium problem is actually a series of $T$ nested equilibrium problems, due to the definition of the policy constraints (\ref{eq:policy_constraints}). If other agents' control variables, $u_t^{-i}$, $t>s$, are not treated as decision variables in \cref{eq:agent_fb_optimization}, and the constraints (\ref{eq:policy_constraints}) are ignored, the game turns into a static trajectory-level game. Such games, referred to as Open-loop Dynamic games \cite{basar1999dynamic}, have dedicated solution methods, e.g., \cite{cleac2019algames}, \cite{di2020local}. 

The nested information pattern arising in the problems (\ref{eq:agent_fb_optimization}) is a fundamental aspect of the method presented in \cref{sec:methods}. The structure of Dynamic Feedback games allows for optimizing over cost-functions and constraints that would otherwise be impossible in standard Open-loop games. Although the nested structure introduced can pose computational difficulties to solve, efficient solvers for approximating equilibria of these games have been implemented, e.g.,  \cite{fridovich2019efficient}. Details on the particular solution approach we use in this work are given in \cref{sec:results}.

%% file: methods.tex
\section{Methods}
\label{sec:methods}

We now present a method which accounts for common forms of uncertainty arising in game-theoretic planning frameworks. In particular, we presume that interaction between agents are only due to collision-avoidance constraints. 

The objective and other constraints imposed upon agents in the scene can often be expressed solely as functions of the private state and control variables for independent agents. Qualitatively different behaviors result from different assignments of responsibility for collision-avoidance. These constraints can be symmetrical, meaning in a pair of agents, both are responsible, or asymmetrical, meaning only one of the two agents is responsible. Asymmetrical situations are useful from a modeling perspective for situations such as when one agent approaches another from behind on the highway, where the rear agent is responsible for avoiding collision. 
Always assigning collision-avoidance responsibility to the ego agent is perhaps the safest option, although this can result in overly-conservative behavior, such as being unable to merge into dense traffic. However, assigning collision-avoidance responsibility to other agents in the scene is risky, since assuming other agents will take responsibility can result in dangerous driving behavior from the ego agent if this assumption is incorrect. 

Our method uses asymmetrical constraints to account for uncertainty the ego vehicle has about the other agents in the scene. In order to do this, we introduce a way to interpolate constraint responsibility between two agents.

Given multiple hypotheses regarding the intentions of each non-ego agent in the scene, we propose introducing a copy of the corresponding agent for each hypothesis. These replicas are endowed with objective and constraint terms of the form 
\cref{eq:agent_fb_optimization},
which reflects the hypothesized intentions. We associate a probability of occurrence with each hypothesis replica. Then, for hypotheses with low probability, we shift collision-avoidance constraint responsibility to the replica agent, allowing the ego to (mostly) ignore the replica. For hypotheses with high probability, we shift constraint responsibility to the ego vehicle, allowing the replica to (mostly) ignore the ego vehicle. A continuum of behaviors in between these two extremes is achieved for intermediate probabilities. 

\subsection{Interpolating Responsibility of Collision-Avoidance}
\label{subsec:constraints}

The interpolation between collision-avoidance constraint responsibility is made possible by the properties of the Dynamic Feedback Game introduced in \cref{subsec:dynamic_feedback}. 

Consider two agents in a Dynamic Feedback Game, one of them being the ego agent. Denote these two agents as \player{1} (ego) and \player{2}. We denote the \emph{independent} objective function of both players to take the form in \cref{eq:obj}, but only depend on the controls and state variables associated with their self:

\begin{subequations}
\label{eq:independent_objectives}
\begin{align}
    L^1(x_0,u_0,...,x_T) := \sum_{t=0}^{T-1} l^1(x_t^1,u_t^1) + l_T^1(x_T^1). \\
    L^2(x_0,u_0,...,x_T) := \sum_{t=0}^{T-1} l^2(x_t^2,u_t^2) + l_T^2(x_T^2). 
\end{align}
\end{subequations}

We add a copy of the objective of \player{2} to the objective of \player{1}, weighted by the odds (corresponding to probability $p$) of this hypothesis for \player{2}. Specifically, the complete objective function for the ego agent is 
\begin{equation}
\begin{aligned}
\label{eq:polite_objective}
    \bar{L}^1(x_0,u_0,...,x_T; p) &:= \\
    L^1(x_0,u_0,...,&x_T) + \frac{p}{1-p} L^2(x_0,u_0,...,x_T).
\end{aligned}
\end{equation}
We refer to the added term in \cref{eq:polite_objective} as a ``politeness'' term. 

The constraints imposed on \player{1} in \crefrange{eq:eq_t}{eq:ineq_T} only depend on its own state and control variables. Conversely, the constraints for \player{2} include the collision-avoidance constraints between \player{1} and \player{2}. These are represented in the constraints $g^2_t(x_t,u_t^2) \geq 0$ and $g^2_T(x_T) \geq 0$. In other words, we only explicitly require \player{2} to account for the collision-avoidance constraints between \player{1} and \player{2}.

Although \player{1} does not account for the collision-avoidance constraints itself, \player{1} can take effective ownership of the constraints for large values of $p$. When $p\to1$, the right-hand side of \cref{eq:polite_objective} dominates \player{1}'s independent objective. This places a very large penalty on sub-optimal values of \player{2}'s objective. This can effectively be viewed as imposing a constraint on \player{1} that \player{2}'s objective term is minimized with respect to the decision variables of \player{1}. 

In this limiting case, \player{1} does its best to ensure that \player{2} does not have to incur any unnecessary cost to optimize its objective or satisfy its constraints, including the collision-avoidance constraints it is responsible for satisfying. Therefore, \player{1} prioritizes staying clear of \player{2}'s desired trajectory, so that \player{2} doesn't have to exert effort to avoid \player{1}. In the other limit, when $p\to0$, the right-hand side of \cref{eq:polite_objective} vanishes, and \player{1} ignores all notions of collision-avoidance, leaving \player{2} to be responsible. For intermediate values of $p$, the responsibility of collision avoidance is shared between the two agents. 

\subsection{Handling Unknown Intentions} 
\label{subsec:unknown_intentions}

Given the ability to interpolate between the collision-avoidance constraint ownership between agents, we propose using the probability of existence associated with multiple hypothesized agents as the interpolating factor appearing in \cref{eq:polite_objective}. 

In particular, assume again that the ego-vehicle is indexed as agent $1$, and all other agents in the scene correspond to indices $2$ through $N$. For each of the other agents, there are $K^i$ hypotheses for the potential intentions of each agent $i$. Assume there is a known categorical belief distribution $Q^i$ over these $K^i$ hypotheses. $Q^i$ is comprised of probabilities $\{p^{i,1}, ..., p^{i,K^i}\}$, where $0<p^{i,k}<1$ and $\sum_{k=1}^{K^i} p^{i,k} = 1$.

We construct a Dynamic Feedback game from the set of hypotheses in the following way. For each hypothesis $k$ regarding agent $i$, we form a replica agent, denoted $P^{i,k}$. Let the independent state of agent $P^{i,k}$ at time-step $t$ be denoted by $x_t^{i,k}$, and the corresponding control variable by $u_t^{i,k}$. The total number of agents (including replicas) is now given by $\hat{N} = 1+ \sum_{i=2}^N K^i$. Let $x_t$ and $u_t$ denote the vector of states and controls for all $\hat{N}$ agents, i.e. $x_t := (x_t^1, x_t^{2,1}, \dots, x_t^{2,K^2}, \dots, x_t^{N,K^N})$, and $u_t$ is defined analogously. 

Each replica agent $P^{i,k}$ is assigned an objective $L^{i,k}$ of the form in \cref{eq:independent_objectives}, which only depends on the states $x_t^{i,k}$ and controls $u_t^{i,k}$. The constraints associated with each agent include collision-avoidance constraints with the ego agent. Here, we assume that non-ego agents in the scene do not interact (i.e., do not avoid collision) with one another. This circumvents the difficulty of, for example, choosing which replica of \player{3} the various replicas of \player{2} interact with. This assumption is easily relaxed in practice, by simply choosing reasonable assignments, such as that $P^{2,1}$ interacts with $P^{3,2}$ and not $P^{3,1}$. In full generality, replicas can be made for every combination of potential non-ego intentions.

The complete objective for the ego agent is given by 
\begin{equation}
\label{eq:full_method_objecdtive}
\begin{aligned}
    \bar{L}^1(x_0,u_0,...,x_T) &:= L^1(x_0,u_0,...,x_T)\\  + &\sum_{i=2}^N\sum_{k=1}^{K^i} \frac{p^{i,k}}{1-p^{i,k}}L^{i,k}(x_0,u_0,...,x_T)
\end{aligned}
\end{equation}

Here, $L^1$ denotes the independent objective as given in \cref{eq:independent_objectives}. The constraints for the ego agent only depend the on the state and control variables  associated to the ego agent (with the exception of the dynamic constraints).

When the constraints and objectives of the ego agent and all replica agents are solved in a Dynamic Feedback Game as in \cref{subsec:dynamic_feedback}, the same notions of constraint ownership described in \cref{subsec:constraints} apply. If any given hypothesis associated with a particular agent has high probability, the politeness term associated with that hypothesis will force the ego agent to take constraint ownership.

%% file: results.tex
\section{Results}
\label{sec:results}

We demonstrate this approach in several different traffic scenarios. 
In each, the ego agent maintains multiple hypotheses about one or more other agents in the scene. By varying its belief about which version of the various agents will realize, the ego generates a spectrum of maneuvers, all of which are game-theoretic equilibria of the game posed in \cref{subsec:unknown_intentions}. For simplicity, we presume that all agents follow a linear driving model with independent lateral and longitudinal accelerations. Nothing in our method precludes the use of more expressive dynamics models, however.

\subsection{Passing Slow-Moving Traffic}
\label{subsec:dense_traffic}

Consider a common situation occurring on highways or roads with two or more lanes in each direction, in which there is slow-moving traffic in one lane, and the ego vehicle alone occupies the other lane. This is the situation depicted in \cref{fig:front}. Although there are no vehicles in front of the ego vehicle forcing it to slow down, it is unsafe to travel at high speeds past slow-moving traffic. Our method allows a natural way to achieve safe driving behavior that is reflective of the probability that one of the slow-moving vehicles will turn into the lane of the ego vehicle. 

Specifically, the two hypotheses considered in this example correspond to \player{2} lane changing vs. staying to the right. In terms of the Dynamic Feedback Game, we represent these hypotheses with the terminal constraints on the lateral position of the red agent. Associated with these two hypotheses is the distribution $Q := \{p, 1-p\}$, where $p$ represents the probability of the  lane-change hypothesis. The independent objectives for all agents are the sums of quadratic costs on their private control inputs (accelerations) and quadratic costs on desired speed. The ego agent additionally minimizes the odds-weighted independent objective of the two red agent replicas, as described in \cref{subsec:unknown_intentions}. Because the non-lane-changing version of the red agent does not interact with the ego agent, the independent objective corresponding to that hypothesis can  be ignored. 

As demonstrated in \cref{fig:front}, varying the probability $p$ naturally produces a range of behaviors of the ego agent, ranging from slowing down in full-anticipation of a lane change, to speeding by, as if the ego vehicle is ignoring the possibility of a lane change.

\subsection{Lane Change: Coupled Predictions}
\label{subsec:lane_change}

\begin{figure}
    \centering
    \includegraphics[width=\columnwidth]{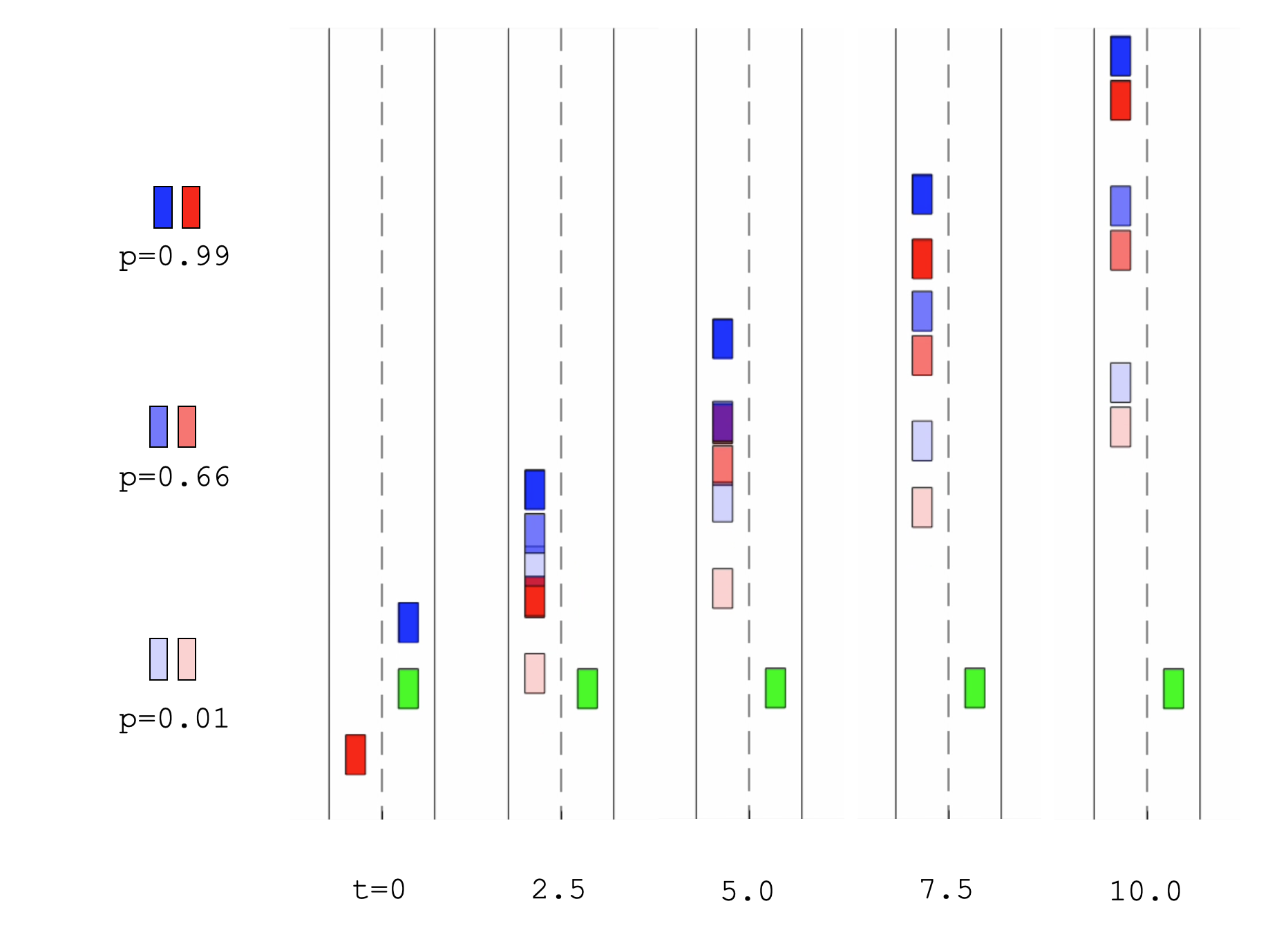}
    \caption{The ego (blue) agent is attempting to change lanes in front of a fast-approaching vehicle (red) in the target lane. The ego agent maintains two hypotheses about the  speed of the red agent. A belief probability $p$ is placed on the hypothesis that the red-vehicle is traveling very fast, as opposed to moderately fast. Varying the probability $p$ results in a spectrum of behaviors for both agents. }
    \label{fig:lane_change}
\end{figure}

\begin{figure}
    \centering
    \includegraphics[width=\columnwidth]{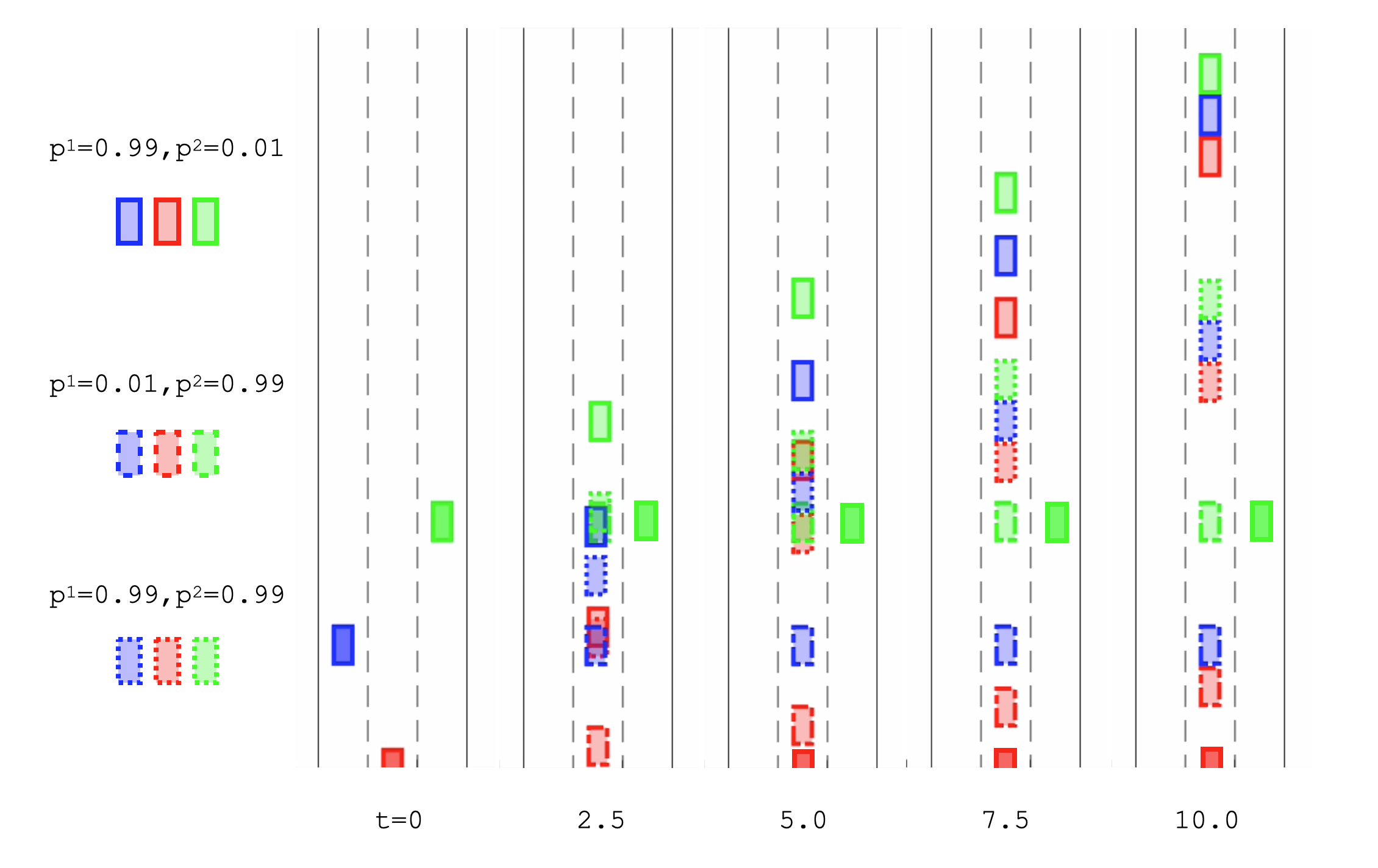}
    \caption{The ego (blue) agent is attempting to change lanes in front of the red agent. The ego agent maintains two hypotheses about the speed of the red agent---(1) that it is traveling fast ($p^1\approx1)$ and (2) that it is traveling at the same speed as ego ($p^1\approx0$). The ego agent additionally maintains two hypotheses about the green agent---(1) that it may also change lanes into the middle lane ($p^2\approx1)$, or (2) may not ($p^2\approx0$). In both hypotheses, the green agent is assumed to travel at the same speed as ego. By varying the belief associated with these independent hypotheses, various behaviors emerge. }
    \label{fig:double_lane_change}
\end{figure}

Consider a similar situation, except now the ego vehicle is on the right and attempting to change lanes, and it is uncertain about the speed of an agent approaching in the left lane. This situation is depicted in \cref{fig:lane_change}. The ego agent again considers two hypotheses regarding the approaching agent, associated with different speeds, with $p$ denoting the probability of that the red agent approaches at the higher speed. Unlike the example in \cref{subsec:dense_traffic}, the hypothesis associated with a lower speed of the approaching agent still requires the ego vehicle to apply some otherwise-non-optimal acceleration. As in the previous example, interpolating $p$ from $0$ to $1$ results in behaviors ranging from respecting the slower-moving hypothesis to respecting the faster-moving hypothesis.  



For intermediate values of $p$, although the odds factor for the slower-moving hypothesis does not vanish, the independent objective of the corresponding replica is not affected by the ego agent's decision variables. This is because since the ego agent is also considering the fast-moving hypothesis, it does not obstruct the agent in the slower-moving hypothesis. The result of this is that the equilibria associated with intermediate values of $p$ will not linearly interpolate between the behaviors associated with $p\approx0$ and $p\approx1$. This is seen in \cref{fig:lane_change}, in which $p=0.66$ is roughly associated with a linear interpolation between the two other behaviors, as opposed to $p=0.5$ as one might expect.  

\subsection{Double Lane-Change: Multiple Independent Hypotheses}

This situation demonstrates the ability of our method to handle hypotheses associated with multiple agents in the scene. As depicted in \cref{fig:double_lane_change}, the ego agent is attempting to lane-change in front of the red agent. There is another (green) vehicle which may also change lanes into the target lane. The ego agent maintains two hypotheses about the speed of the red agent, one being that it is traveling at the same speed as ego, and the other that it is traveling much faster. The ego agent also maintains two hypotheses regarding the green agent, one in which it changes lane and one in which it doesn't. We associate a probability $p^1$ with the hypothesis that the red agent is traveling fast, and $p^2$ with the hypothesis that the green agent will change lanes. 

The behaviors generated for various values of probabilities $p^1$ and $p^2$ are shown in \cref{fig:double_lane_change}. As in previous examples, when the probability of a hypothesis is high, the ego agent takes ownership of collision-avoidance with the corresponding replica, and when the probability is low, the hypothesis is effectively ignored. An interesting aspect of this example is the case in which both $p^1\to1$ and $p^2\to1$. In this case, the ego agent is conflicted between traveling fast to be polite with respect to the approaching red agent, and traveling slow to allow the green agent to also change lanes. The compromise is to travel at a moderate speed, trading off the preference of both agents. From a practical perspective, this can be problematic. Even if the ego agent is certain that the red agent is moving fast, taking ownership of collision-avoidance with respect to the green agent should take precedence over avoiding collision with the red agent. This behavior can be achieved by simply requiring $p^1<p^2$ by a sufficient margin. In this perspective, the values $p^1$ and $p^2$ would not be interpreted directly as probability of hypothesis, but simply as \say{politeness} parameters. 

\subsection{Computation}
The method we used to compute solutions in the above examples is a version of the method described in \cite{fridovich2019efficient}, modified to account for constraints. In particular, we implemented a method analogous to Sequential Quadratic Programming to jointly solve the first-order necessary conditions corresponding to the game. Linear-Quadratic (LQ) problems formed at each major iteration are solved using an Active-Set (AS) approach analogous to those used in quadratic programming \cite{nocedal2006numerical}. The equality-constrained LQ games to be solved in each minor iteration of the AS approach are solved using a method analogous to \cite{laine2019efficient}, adapted to the game setting. Although these algorithmic components are complex and bear discussion in their own right, due to space limitations we defer a more detailed treatment of algorithms for feedback dynamic games to a separate publication.

\begin{table}[]
    \centering
    \vspace{.2cm}
    \begin{tabular}{|c|c|c|c|c|}
        \hline 
         Example & $p$ & Total (s) & Data (s)  & LQ Iters  \\ \hline 
         A & $p= $0.01 & 2.15 & 0.77 & 7 \\
         A & $p= $0.50 & 1.99 & 0.73  & 7 \\
         A & $p= $0.99 & 3.12 & 2.03  & 11 \\
         \hline 
         B & $p= $0.01 & 2.26 & 0.73  & 8 \\
         B & $p= $0.66 & 19.08 & 5.24 & 70 \\
         B & $p= $0.99 & 2.35 & 0.71 & 9 \\
         \hline 
         C & $p^1= $0.99, $p^2= $0.01 & 1.92 & 0.70 & 7 \\
         C & $p^1= $0.01, $p^2= $0.99 & 3.43 & 1.04 & 13 \\
         C & $p^1= $0.99, $p^2= $0.99 & 30.76 & 6.62 & 113 \\
         \hline
    \end{tabular}
    \caption{Solve times and iterations for all examples. Example IDs are according to subsections of \cref{sec:results}. The information under ``Total'' refers to total solve time. ``Data'' refers to the portion of total time spent evaluating problem data such as function gradients and Hessians. ``LQ Iters'' refer to the number of solves of equality-constrained LQ-games in minor iterations of the solver. Both instances with exceptionally long solve-times had very-large numbers of active-constraints, necessitating many iterations due to poor initializations. }
    \label{tab:times}
\end{table}

Run-time was not a major concern for the purposes of this work. For ease of prototyping, we implemented our method in MATLAB\textsuperscript{\textregistered}. All examples were solved with 50 discrete knot points, with solve times listed in \cref{tab:times}. Although there are many known inefficiencies in our implementation, we strongly believe that the method presented in this work is amenable to real-time computation ($\SI{0.5}{\second}$), if implemented efficiently. For example, proper initializations of the solver can avoid unnecessary iterations, especially when using an active-set method as we do here. Exploiting linearity can further avoid unnecessary computations of gradients or Hessians, and utilizing parallelization to compute problem data could result in large savings. Parallelization could also be utilized in LQ solves as demonstrated in \cite{laine2019parallel}. Finally, implementation in a strongly typed, compiled language such as C++ or Julia would result in major speedups as well. 
Code for reproducing all examples will be released at \href{https://www.github.com/4estlaine/gfne}{https://www.github.com/4estlaine/gfne}.

%% file: conclusion.tex
\section{Conclusion}
\label{sec:conclusion}

This paper accounts for categorical uncertainty in the intentions of other players by forming replicas of those players, and assigning collision-avoidance responsibility according to the probability associated to each replica. We present the method in full, and show several examples of its expressive ability to handle uncertain intentions in the context of autonomous driving. 

Future work will focus on a real-time implementation of our method, and will investigate the application of this method to receding horizon, model-predictive control settings. It may also be important to study our method in contexts where interactions among non-ego agents cannot be ignored. Finally, methods to estimate agents' constraints should be investigated. Preliminary work in the context of optimal control is ongoing, e.g., \cite{scobee2019maximum, vazquez2020maximum, chou2020learning}, but should be extended to the game setting. 

%% file: books.bib
@book{basar1999dynamic,
  title={{Dynamic Noncooperative Game Theory}},
  author={Ba\c{s}ar, Tamer and Olsder, Geert Jan},
  volume={23},
  year={1999},
  publisher={SIAM},
  edition={2nd}
}

@book{nocedal2006numerical,
  title={Numerical optimization},
  author={Nocedal, Jorge and Wright, Stephen},
  year={2006},
  publisher={Springer Science \& Business Media}
}

@article{fridovich2019efficient,
  title={Efficient Iterative Linear-Quadratic Approximations for Nonlinear Multi-Player General-Sum Differential Games},
  author={Fridovich-Keil, David and Ratner, Ellis and Dragan, Anca D and Tomlin, Claire J},
  journal={arXiv preprint arXiv:1909.04694},
  year={2019}
}

@inproceedings{sadigh2016planning,
  title={Planning for autonomous cars that leverage effects on human actions.},
  author={Sadigh, Dorsa and Sastry, Shankar and Seshia, Sanjit A and Dragan, Anca D},
  booktitle={Robotics: Science \& Systems},
  volume={2},
  year={2016},
  organization={Ann Arbor, MI, USA}
}

@article{Sadigh2016information,
  author = {Sadigh, Dorsa and Sastry, Shankar and Seshia, Sanjit A. and Dragan, Anca D.},
  journal = {International Conference on Intelligent Robots and Systems (IROS)},
  title = {{Information Gathering Actions over Human Internal State}},
  year = {2016}
}

@article{Aoude2013probabilistically,
 author = {Aoude, Georges S. and Luders, Brandon D. and Joseph, Joshua M. and Roy, Nicholas and How, Jonathan P.},
 title = {Probabilistically Safe Motion Planning to Avoid Dynamic Obstacles with Uncertain Motion Patterns},
 journal = {Autonomous Robots},
 volume = {35},
 number = {1},
 year = {2013},
 pages = {51--76},
 numpages = {26},
 url = {http://dx.doi.org/10.1007/s10514-013-9334-3},
 doi = {10.1007/s10514-013-9334-3},
}

@inproceedings{ziebart2009planning,
  title={Planning-based prediction for pedestrians},
  author={Ziebart, Brian D. and Ratliff, Nathan and Gallagher, Garratt and Mertz, Christoph and Peterson, Kevin and Bagnell, J. Andrew and Hebert, Martial and Dey, Anind K and Srinivasa, Siddhartha},
  booktitle={International Conference on Intelligent Robots and Systems (IROS)},
  pages={3931--3936},
  year={2009},
  organization={IEEE}
}


%% file: papers.bib
@article{nishimura2020rat,
  title={RAT iLQR: A Risk Auto-Tuning Controller to Optimally Account for Stochastic Model Mismatch},
  author={Nishimura, Haruki and Mehr, Negar and Gaidon, Adrien and Schwager, Mac},
  journal={arXiv preprint arXiv:2010.08174},
  year={2020}
}

@article{hardy2013contingency,
  title={Contingency planning over probabilistic obstacle predictions for autonomous road vehicles},
  author={Hardy, Jason and Campbell, Mark},
  journal={IEEE Transactions on Robotics},
  volume={29},
  number={4},
  pages={913--929},
  year={2013},
  publisher={IEEE}
}

@article{van2015distributionally,
  title={Distributionally robust control of constrained stochastic systems},
  author={Van Parys, Bart PG and Kuhn, Daniel and Goulart, Paul J and Morari, Manfred},
  journal={IEEE Transactions on Automatic Control},
  volume={61},
  number={2},
  pages={430--442},
  year={2015},
  publisher={IEEE}
}

@article{scobee2019maximum,
  title={Maximum likelihood constraint inference for inverse reinforcement learning},
  author={Scobee, Dexter RR and Sastry, S Shankar},
  journal={arXiv preprint arXiv:1909.05477},
  year={2019}
}

@inproceedings{vazquez2020maximum,
  title={Maximum Causal Entropy Specification Inference from Demonstrations},
  author={Vazquez-Chanlatte, Marcell and Seshia, Sanjit A},
  booktitle={International Conference on Computer Aided Verification},
  pages={255--278},
  year={2020},
  organization={Springer}
}

@article{chou2020learning,
  title={Learning constraints from locally-optimal demonstrations under cost function uncertainty},
  author={Chou, Glen and Ozay, Necmiye and Berenson, Dmitry},
  journal={IEEE Robotics and Automation Letters},
  volume={5},
  number={2},
  pages={3682--3690},
  year={2020},
  publisher={IEEE}
}

@INPROCEEDINGS{laine2019parallel,
  author={F. {Laine} and C. {Tomlin}},
  booktitle={2019 IEEE 58th Conference on Decision and Control (CDC)}, 
  title={{Parallelizing LQR Computation Through Endpoint-Explicit Riccati Recursion}}, 
  year={2019},
  volume={},
  number={},
  pages={1395-1402},
  doi={10.1109/CDC40024.2019.9029974}}

@inproceedings{laine2019efficient,
  title={{Efficient computation of feedback control for equality-constrained LQR}},
  author={Laine, Forrest and Tomlin, Claire},
  booktitle={2019 International Conference on Robotics and Automation (ICRA)},
  pages={6748--6754},
  year={2019},
  organization={IEEE}
}

@inproceedings{di2020local,
  title={Local First-Order Algorithms for Constrained Nonlinear Dynamic Games},
  author={Di, Bolei and Lamperski, Andrew},
  booktitle={2020 American Control Conference (ACC)},
  pages={5358--5363},
  year={2020},
  organization={IEEE}
}

@article{facchinei2010generalized,
  title={Generalized Nash equilibrium problems},
  author={Facchinei, Francisco and Kanzow, Christian},
  journal={Annals of Operations Research},
  volume={175},
  number={1},
  pages={177--211},
  year={2010},
  publisher={Springer}
}

@inproceedings{xu2014motion,
  title={Motion planning under uncertainty for on-road autonomous driving},
  author={Xu, Wenda and Pan, Jia and Wei, Junqing and Dolan, John M},
  booktitle={2014 IEEE International Conference on Robotics and Automation (ICRA)},
  pages={2507--2512},
  year={2014},
  organization={IEEE}
}

@article{facchinei2009generalized,
  title={Generalized Nash equilibrium problems and Newton methods},
  author={Facchinei, Francisco and Fischer, Andreas and Piccialli, Veronica},
  journal={Mathematical Programming},
  volume={117},
  number={1-2},
  pages={163--194},
  year={2009},
  publisher={Springer}
}

@article{schwarting2019stochastic,
  title={Stochastic Dynamic Games in Belief Space},
  author={Schwarting, Wilko and Pierson, Alyssa and Karaman, Sertac and Rus, Daniela},
  journal={arXiv preprint arXiv:1909.06963},
  year={2019}
}

@inproceedings{di2019newton,
  title={Newton’s Method and Differential Dynamic Programming for Unconstrained Nonlinear Dynamic Games},
  author={Di, Bolei and Lamperski, Andrew},
  booktitle={2019 IEEE 58th Conference on Decision and Control (CDC)},
  pages={4073--4078},
  year={2019},
  organization={IEEE}
}

@article{cleac2019algames,
  title={ALGAMES: A Fast Solver for Constrained Dynamic Games},
  author={Cleac'h, Simon Le and Schwager, Mac and Manchester, Zachary},
  journal={arXiv preprint arXiv:1910.09713},
  year={2019}
}

@article{peters2020inference,
  title={Inference-Based Strategy Alignment for General-Sum Differential Games},
  author={Peters, Lasse and Fridovich-Keil, David and Tomlin, Claire J and Sunberg, Zachary N},
  journal={arXiv preprint arXiv:2002.04354},
  year={2020}
}
